\documentclass{article}
\usepackage{spconf,amsmath,graphicx,hyperref}

\usepackage[utf8]{inputenc}
\usepackage[T1]{fontenc}
\usepackage{microtype}
\usepackage{nicefrac}
\usepackage{textcomp}

\usepackage{amsmath,amssymb,amsfonts}
\usepackage{mathtools}
\usepackage{scalerel}

\usepackage{graphicx}
\usepackage{booktabs}
\usepackage{multirow}
\usepackage{longtable}
\usepackage{caption}
\usepackage{subcaption}
\usepackage{wrapfig}
\usepackage{float}
\usepackage{placeins}
\usepackage{eqparbox}
\usepackage{makecell}

\usepackage{algorithmic}

\usepackage{enumitem}
\usepackage{comment}
\usepackage{lipsum}

\usepackage{xcolor}
\usepackage{url}
\usepackage{hyperref}
\usepackage[capitalize]{cleveref}

\Crefname{Chapter}{Chap.}{Chaps.}
\Crefname{Section}{Sec.}{Secs.}
\Crefname{Figure}{Fig.}{Figs.}

\usepackage{glossaries}

\glsdisablehyper

\newacronym{asr}{ASR}{automatic speech recognition}
\newacronym{asa}{ASA}{automatic speaking assessment}
\newacronym{cefr}{CEFR}{Common European Framework of Reference for Languages}
\newacronym[plural=CIs,firstplural=confidence intervals (CIs)]
{ci}{CI}{confidence interval}
\newacronym{ctc}{CTC}{connectionist temporal classification}
\newacronym{fh}{FH}{factored hybrid}

\newacronym{gop}{GoP}{Goodness-of-Pronunciation}
\newacronym{vad}{VAD}{Voice Activity Detection}
\newacronym{hmm}{HMM}{Hidden Markov Model}
\newacronym{l2}{L2}{second language}
\newacronym{ssl}{SSM}{self-supervised speech model}
\newacronym{svd}{SVD}{singular value decomposition}
\newacronym{wer}{WER}{word error rate}
\newacronym{dtw}{DTW}{dynamic time warping}

\newacronym{sandi}{S\&I}{Speak and Improve Corpus 2025}
\newacronym{ume}{UME-ERJ}{English Read by Japanese Students dataset}
\newacronym{lbs}{LBS-4h}{4h subset of LibriSpeech 960h}

\def\pn{$\mathcal{P}_{\scaleto{\text{nat}}{3.5pt}}$}
\def\pl{$\mathcal{P}^{r}_{\scaleto{\text{L}2}{3.5pt}}$}

\def\mnative{$M_{\scaleto{\text{nat}}{3.5pt}}$}
\def\mnativebar{$\bar{M}_{\scaleto{\text{nat}}{3.5pt}}$}
\def\ml{$M^{r}_{\scaleto{\text{L}2}{3.5pt}}$}

\def\cnative{$C_{\scaleto{\text{nat}}{3.5pt}}$}
\def\cl{$C^{r}_{\scaleto{\text{L}2}{3.5pt}}$}

\def\cnativestar{$C^{\star}_{\scaleto{\text{nat}}{3.5pt}}$}
\def\clstar{$C^{r,\star}_{\scaleto{\text{L}2}{3.5pt}}$}

\def\cnativestarp{$C^{\star}_{\scaleto{\text{nat},p}{3.5pt}}$}
\def\clstarp{$C^{r,\star}_{\scaleto{\text{L2,p}}{3.5pt}}$}

\def\pc{phone-class}
\def\PC{Phone-Class}
\def\cu{center-unit}
\def\fu{full-unit}

\makeatletter
\renewcommand{\section}{\@startsection
	{section}%
	{1}%
	{}%
	{-0.9\baselineskip}%
	{0.1\baselineskip}%
	{}}%
\renewcommand{\subsection}{\@startsection
	{subsection}%
	{2}%
	{}%
	{-0.5\baselineskip}%
	{0.3\baselineskip}%
	{}}%
\renewcommand{\subsubsection}{\@startsection
	{subsubsection}%
	{3}%
	{}%
	{-0.3\baselineskip}%
	{0.1\baselineskip}%
	{}}%
\g@addto@macro\normalsize{%
	\setlength\abovedisplayskip{5pt plus 2pt minus 2pt}
	\setlength\belowdisplayskip{5pt plus 2pt minus 2pt}
	\setlength\abovedisplayshortskip{4pt plus 2pt minus 2pt}
	\setlength\belowdisplayshortskip{4pt plus 2pt minus 2pt}
}
\makeatother

\title{A Native-Reference Phone-Class Geometry for Second-Language Pronunciation Analysis}
\name{Tina Raissi \quad Nhan Phan \quad Chenxiao Wang \quad Mikko Kurimo}
\address{Department of Information and Communications Engineering,
	Aalto University,
	Espoo, Finland \\
	\texttt{\{firstname\}.\{lastname\}@aalto.fi}}

\begin{document}
\ninept
\maketitle
\begin{abstract}
Automatic speaking assessment systems can provide holistic proficiency scores, but often lack interpretable measures that characterize pronunciation quality.
We propose a native-reference \pc{} geometry for measuring \gls{l2} pronunciation deviation without requiring pronunciation labels, read-aloud prompts, or matched
recordings of the same text from native and \gls{l2} speakers.
Given a native speech corpus, we average frame-level self-supervised representations for each context-dependent \pc{} and use \gls{svd} to derive a compact native-reference coordinate system.
For each \gls{l2} utterance, we compute the corresponding averages and project them into the native-reference space.
We then demonstrate that the distances between \gls{l2} and native-reference coordinates for matched \pc es show consistent negative correlations with holistic speaking proficiency on the Dev subset of the \glsentrylong{sandi} (Spearman's $\rho\!=\!-0.53$) and with pronunciation quality on the learner subset of the \glsentrylong{ume} ($\rho\!=\!-0.34$).
These findings suggest that the proposed geometry captures acoustic-phonetic information relevant for proficiency rating while remaining applicable to spontaneous \gls{l2} speech without matched native recordings.

\end{abstract}
\section{Introduction}
\label{sec:intro}
\glsreset{l2}
\glsreset{sandi}
\glsreset{l2}
\glsreset{ume}
\glsreset{svd}

\Gls{asa} aims to evaluate \gls{l2} speech and has growing applications in language proficiency rating and learning.
Existing systems combine information from several dimensions, including fluency, pronunciation, lexical use, and grammar, to estimate holistic speaking proficiency~\cite{chen_speechrater_2018, al-ghezi_ASA2023a}. 
Among these dimensions, pronunciation is usually the most challenging to address due to fine-grained acoustic variations over time.
While holistic scores summarize overall speaking ability, understanding the relationship between individual speech properties and these scores is important for interpreting assessment outcomes and developing targeted feedback for language learners.
Acoustic representations corresponding to phonetic segments provide one way of characterizing variation in \gls{l2} speech. 
Previous work has shown that distances derived from phone-specific acoustic representations can contain information associated with overall speaking proficiency, including for spontaneous speech~\cite{kyriakopoulos_phone_distrance_2018}. 
From a speech processing perspective, a key source of information derives from the acoustic realization of words, and more specifically subword units, by \gls{l2} speakers.
This motivates the development of reliable pronunciation assessment models that can capture the relevant acoustic cues while remaining robust to \gls{l2} speech variability.
Methods such as \gls{gop} are well suited to constrained read-aloud settings, where the expected prompt is known and phone-level pronunciation evidence can be derived from acoustic model likelihoods or posterior probabilities. 
Applying such approaches to spontaneous \gls{l2} speech is more challenging, as disfluencies and pronunciation variability can make automatic transcription and phone alignment less reliable, while acoustic models may also be affected by mismatch between their training data and \gls{l2} speech.
This issue is particularly relevant for frame-wise models with blank-based label topologies, such as \gls{ctc}, where peaky output distributions~\cite{bluche2015phd} and the role of the blank label complicate the interpretation of frame-level posterior evidence. 
This motivates a complementary question: rather than deriving pronunciation evidence from likelihood or posterior distributions over phone labels, can we instead analyze the continuous phone representations directly?

Recent research has examined the phonetic information encoded by self-supervised speech models~\cite{pasad2021,yang21c_interspeech,wells2022phonetic,pasad2023comparative}, showing that representations extracted from phone-aligned segments capture context-dependent phone information~\cite{choi2025leveraging,choi2026b} and extend beyond the corresponding phone to reflect broader contextual information~\cite{pasad2024self}.
Nevertheless, these findings do not directly provide an interpretable metric for \gls{l2} pronunciation assessment.
Distance-based approaches have also been explored using dynamic time warping alignments between \gls{l2} and native speech~\cite{bartelds2020new,chernyak2024perceptual,mcintosh2026self}.
Such methods, however, require matched recordings of the same text from
native and \gls{l2} speakers and can be sensitive to disfluencies, which are common in \gls{l2} speech.

We propose a native-reference \pc{} geometry that is applicable to spontaneous \gls{l2} speech and does not require matched linguistic content between native and learner utterances.
Instead of comparing whole utterances, we first construct context-dependent \pc{} \emph{centroids} by averaging frame-level self-supervised representations over their corresponding segments.
By performing a low-rank \gls{svd} of the native centroid matrix, we define a low-dimensional native-reference basis that captures dominant variation among native \pc{} realizations. 
For each \gls{l2} utterance, we project its \pc{} centroids into the native projection basis, yielding L2 \pc{} coordinates.
We then measure the pronunciation deviation as the distance between \gls{l2} and native coordinates for matching \pc es, using different distance measures.
In particular, we do not train a model using the centroids or the distances to predict the pronunciation or proficiency level.
Instead, we evaluate whether the raw native-reference distances are monotonically associated with the available human ratings: holistic speaking proficiency for \gls{sandi} and pronunciation quality for \gls{ume}.

This paper extends our earlier work~\cite{raissi2026native} and further assesses the behavior and validity of our proposed method by evaluating spontaneous speech from \gls{sandi} and read-aloud speech from \gls{ume} across several experimental settings.
Since comparable results are not available in prior work, we designed a series of baseline experiments.
We then test whether the observed association with proficiency (1) is robust to the choice of native-reference corpora across TIMIT, \gls{lbs}, and native subset of \gls{ume}, (2) transfers to the read-aloud \gls{ume} pronunciation setting, and (3) depends on methodological choices such as projection space or alignment quality.
Our native-reference distances show consistent negative correlations with proficiency, suggesting that the proposed geometry captures relevant acoustic information without requiring read-aloud speech or matched recordings of the same
text from native and \gls{l2} speakers.

\section{Proposed Method}
\label{sec:method}

Let $H^{\ell} \in \mathbb{R}^{T \times D}$ denote the sequence of encoder representations extracted from layer $\ell$ of a self-supervised speech model. 
Given the phonetic transcription of a spoken word sequence, we define context-dependent \emph{\pc es} by augmenting each center phone with left context for diphones and both left and right context for triphones.
We denote by $\mathcal{P}$ the set of resulting \pc es.
Given an aligned \pc{} sequence 
$\{a_t\}_{t=1}^T$ obtained by forced alignment, we define $\mu(p)=\frac{1}{|\mathcal{T}_p|}\sum_{t\in\mathcal{T}_p} h_t$ as the \pc{} \emph{centroid} for class $p \in \mathcal{P}$, where $\mathcal{T}_p \coloneqq \{t: a_t = p\}$.
The centroids are obtained by using feature-slicing~\cite{pasad2021} with two averaging strategies. 
In \emph{\cu{}} averaging, only the frames aligned to the center phone are used. 
In \emph{\fu{}} averaging, all frames belonging to the complete context-dependent unit, including its contextual phones, are pooled. 
Each centroid summarizes the typical realization of a phone in a specified phonetic context encoded by the self-supervised representation.
The resulting \pc{} centroids are stacked as the rows of a \emph{centroid matrix} $M\in\mathbb{R}^{|\mathcal{P}|\times D}$.

\subsection{Native-Reference and \PC{} Coordinates}
\label{subsec:codebook}

The native centroid matrix \mnative{} is constructed by considering the set of \pc es derived from the transcriptions of a native speech corpus. 
Since the rows of \mnative{} are pooled across all native utterances, each centroid for a \pc{} $p\in\text{\pn}$ represents a corpus-level estimate rather than a speaker or sentence-specific realization.
We introduce a global native mean vector $\bar{\mu}_{\text{nat}}=\tfrac{1}{\vert\mathcal{\text{\pn}}\vert}\sum_{p\in\text{\pn}}\mu(p)$, and denote \vspace{-0.2cm} 
$$\text{\mnativebar}= \text{\mnative} - \mathbf{1} \bar{\mu}^{\top}_{\text{nat}}$$
as the centered centroid matrix.
In order to obtain a native-reference \pc{} coordinate system, we perform rank-$K$ truncated \gls{svd} of the native centered centroid matrix
\begin{equation}\nonumber \footnotesize
	\text{\mnativebar}\approx U\Sigma V^{\top},\quad
	U\in\mathbb{R}^{\vert\text{\pn}\vert\times K},\;\Sigma\in\mathbb{R}^{K\times K},\;V\in\mathbb{R}^{D\times K},
\end{equation}
where $U\Sigma V^{\top}$ is the best possible rank-$K$ linear approximation of the centered native \pc{} mean matrix under Frobenius norm~\cite{eckart1936approximation}.
The \emph{projection basis matrix} $V$ contains orthonormal directions spanning a $K$-dimensional native-reference subspace of the original self-supervised feature space, capturing dominant variation among centered native \pc{} mean vectors.
The first direction $v_1$ captures the largest variation among native \pc{} centroids, while subsequent directions capture the largest remaining orthogonal variations. 
The rank $K$ controls how much of the native acoustic-phonetic variation is retained in the coordinate system.

We now define $\text{\cnative}=U\Sigma\in\mathbb{R}^{\vert\mathcal{\text{\pn}}\vert\times K}$ as the \emph{native-reference coordinate matrix}. 
Its rows contain the coordinates of the \pc{} mean vectors in the basis defined by $\{v_k\}_{k=1}^K$, with each row corresponding to one \pc{} and each column to one \gls{svd} direction. 
While the columns of $V$ define the native-reference basis, the matrix $\text{\cnative}$ contains the coordinates of the native \pc{} centroids in this basis.
For an \gls{l2} utterance $r$ and its self-supervised feature sequence $H_r^{\ell} \in \mathbb{R}^{T_r \times D}$, we build the \gls{l2} centroid matrix $\text{\ml}\in\mathbb{R}^{\vert\mathcal{\text{\pl}}\vert \times D}$ in the same way as for \mnative{} with the difference that now \pl{} is the set \pc es appearing in the transcript of the L2 utterance $r$.
Unlike the native centroid matrix, \ml{} is utterance-level, i.e., it contains only the \pc es observed in the current \gls{l2} utterance. 
The corresponding \emph{\gls{l2} coordinate matrix} is then calculated by projecting \ml{} into the native \gls{svd} subspace $V$, after subtracting the native global mean $\bar{\mu}_{\text{nat}}$, resulting in $\text{\cl} = (\text{\ml} -  \mathbf{1} \bar{\mu}^{\top}_{\text{nat}} ) V$.

	

\subsection{Native-Reference Distance and Evaluation}
\label{subsec:distance}

For each utterance $r$, we restrict the comparison to the \pc es shared by the native and L2 representations, $\mathcal{P}_r^{\star}=\text{\pn}\cap\text{\pl}$, yielding \cnativestar{} and \clstar{}. 
For each shared \pc{} $p$, we measure the deviation by $\delta_r(p)=d_m\!\left(\text{\cnativestarp},\text{\clstarp}\right)$, where $d_m$ is Euclidean, cosine, or diagonal Mahalanobis distance. 
The utterance-level distance aggregates $\delta_r(p)$ over $p\in\mathcal{P}_r^{\star}$. 
This ensures that each comparison is made between the same \pc{} in native and \gls{l2} speech, while using a fixed native mean and basis for all \gls{l2} utterances.
We evaluate the association between the resulting utterance-level distance and proficiency or pronunciation scores using Spearman's $\rho$~\cite{spearman_1904}. 
We also report Pearson's $r$~\cite{pearson_1896} in our final evaluation as a complementary linear association measure.
Since lower distances indicate greater similarity to the native-reference coordinates and higher scores indicate higher proficiency, negative correlations indicate that native-reference distance decreases with proficiency.

\section{Experiments}
\label{sec:exp}
\subsection{Setting}
\label{subsec:setting}
To construct our native-reference coordinates, we selected 4 hours of speech from each of TIMIT~\cite{garofolo1993timit}, LibriSpeech 960h~\cite{povey2015librispeech}, and \glsfirst{ume}~\cite{minematsu2002english}. 
TIMIT comprises 630 speakers across its three subsets. 
For comparability, we sampled 630 speakers from \gls{lbs}, while \gls{ume} contains only 20 speakers.
While all three native-reference corpora contain read speech, they differ in style: \gls{lbs} is derived from audiobooks, \gls{ume} consists of single sentences read under controlled conditions, and TIMIT provides phonetically balanced speech with ground truth phone segmentation. 
For \gls{l2} speech, we use Train and Dev partitions of the \gls{sandi}~\cite{knill25_slate,qian2025speak} together with the learner subset of \gls{ume}.
For the \gls{sandi} Train, only utterances with manual transcriptions are used.
For both learner and native subsets of \gls{ume}, we only considered the \emph{sentences} partition.
The correlation coefficients are computed between our proposed distances and the proficiency scores for \gls{sandi}, and between our proposed distances and the pronunciation scores for \gls{ume}.
For \gls{sandi}, each subset is divided into four proficiency-evaluation tasks. We compute the correlation coefficient for each task and report the average, following common practice.
For \gls{ume}, each sentence is read by several native and learner speakers, and we follow prior work~\cite{mcintosh2026self} for score aggregation.

We selected the best distance measure, encoder layer, and \gls{svd} rank using \gls{sandi} Train and \gls{ume} learner, with \gls{lbs} serving as the native-reference corpus.
We use speech representations from layers 6 and 12 of WavLM Base+ encoder~\cite{chen2022wavlm} for \gls{sandi} and \gls{ume}, respectively.
As the test scenario, we report results on \gls{sandi} dev-\gls{lbs} and learner-native \gls{ume} using the best configuration, consisting of \gls{svd} rank 32, \pc{} (\gls{sandi}: triphone, \gls{ume}: diphone), and averaging strategy (\gls{sandi}: \fu{}, \gls{ume}: \cu{}).
We use the ARPAbet phones and merge stressed phones into a unique class for a total of 39 phones and one silence label.
For all corpora except TIMIT, which provides ground truth alignments, phone segmentations are obtained by forced alignment using a Conformer-based factored \gls{hmm}~\cite{raissi24_interspeech} trained on the small Loquacious subset~\cite{parcollet25_interspeech}.
For our baselines using \gls{vad}, speech regions are obtained by frame-level majority voting over three systems (Pyannote~\cite{bredin2020pyannote}, Silero VAD~\cite{silero2021vad}, and FireRedVAD~\cite{xu2026fireredasr2s}), after which pauses shorter than 25 milliseconds are merged into the surrounding speech.
We used Lhotse~\cite{zelasko2021lhotse} for data preparation and RASR~\cite{rybach2011rasr} for forced alignment. 
We provide the source code of our experiments~\cite{raissi2026code}.

\subsection{Baselines}
\label{subsec:base}
\begin{table}[t]
	\centering
	\footnotesize
	\setlength{\tabcolsep}{0.9em}
	\renewcommand{\arraystretch}{1.05}
	\caption{Non-acoustic baselines. We report Spearman correlation $\rho$ for \gls{sandi} Train, \gls{sandi} Dev, and \gls{ume} learner using \gls{lbs} native-reference corpus. 
		Higher positive correlations indicate better agreement with proficiency level.
        $\mathcal{P}^\star$ denotes the matched context-dependent \pc es between native and \gls{l2} coordinates in our proposed method.}
	\label{tab:non_acoustic_baselines}
	\begin{tabular}{| l | l | c | c | c |}
		\hline
		\multirow{2}{*}{\textbf{Source}} & \multirow{2}{*}{\textbf{Feature}} & \multicolumn{2}{c|}{\textbf{\gls{sandi}}} & \multirow{2}{*}{\textbf{\gls{ume}}} \\
		\cline{3-4}
		& & \textbf{Train} & \textbf{Dev} & \\
		\hline
		
		\multirow{2}{*}{Raw} & \# Words & $0.55$ & $0.57$ & $-0.20$ \\
		\cline{2-5}
		& Speaking rate & $0.54$ & $0.52$ &$ \phantom{-}0.13$ \\
		\hline
		
		\multirow{2}{*}{\shortstack[l]{$\mathcal{P}^\star$}} 
		& Diphones & $0.60$ & $0.62$ & $-0.23$ \\
		\cline{2-5}
		& Triphones & $0.60$ & $0.62$ & $-0.22$ \\
		\hline
	\end{tabular}
\end{table}
\begin{table}[t]
	\centering
	\footnotesize
	\setlength{\tabcolsep}{1em}
	\renewcommand{\arraystretch}{1.05}
	\caption{Acoustic baselines for similar evaluations and datasets as in \cref{tab:non_acoustic_baselines}.
		Here, lower negative correlations indicate better agreement with proficiency.
		The silence ratio is calculated by considering both voice activity detection~(VAD) and silence segments from the alignment.
		Encoder averaging and \gls{dtw} apply only to \gls{ume}, which provides parallel learner–native recordings.}
	\label{tab:acoustic_baselines}
	\begin{tabular}{| l | l | c | c | c |}
		\hline
		\multirow{2}{*}{\textbf{Source}} & \multirow{2}{*}{\textbf{Feature}} & \multicolumn{2}{c|}{\textbf{\gls{sandi} }} & \multirow{2}{*}{\textbf{\gls{ume}}} \\
		\cline{3-4}
		& & \textbf{Train} & \textbf{Dev} & \\
		\hline
		
		\multirow{2}{*}{Silence} & VAD & $-0.36$ & $-0.41$ & $\phantom{-}0.10$ \\
		\cline{2-5}
		& Alignment & $-0.60$ & $-0.57$ & $-0.07$ \\
		\hline
		
		ASR & WER & $-0.22$ & $-0.20$ & $-0.24$ \\
		\hline
		
		Raw & Mean & \multicolumn{2}{c|}{\multirow{2}{*}{N/A}} & $-0.20$ \\
		\cline{2-2} \cline{5-5}
		Encoder & DTW & \multicolumn{2}{c|}{} & $-0.50$ \\
		\hline
	\end{tabular}
    \vspace{-0.25cm}
\end{table}

We first report baseline controls that may correlate with speaker proficiency but do not rely on acoustic information or distance measures.
\Cref{tab:non_acoustic_baselines} shows that, on \gls{sandi}, word count, speaking rate, and matched \pc{}
coverage $\vert\mathcal{P}_r^\star\vert$ correlate strongly with holistic scores ($\rho\approx0.5$-$0.6$).
This is consistent with spontaneous
speech where higher-proficiency speakers produce more fluent and longer
responses, while using a broader vocabulary range, as reflected in greater matched \pc{} coverage.
 In contrast, the same features show weak or negative correlation on \gls{ume}, which consists of read-aloud speech and is rated specifically for pronunciation quality.
Thus, these features provide important information for interpreting whether native-reference distances capture acoustic variation beyond phonetic coverage.
In \cref{tab:acoustic_baselines}, we also report acoustic baselines capturing fluency, \gls{asr} accuracy, and \gls {dtw}-based acoustic similarity between speakers producing matched sentences.
On \gls{sandi}, silence shows a strong negative correlation, especially when derived from an alignment ($\rho\approx-0.60$), confirming that the proficiency level is strongly
affected by pauses. 
In contrast, these features are weak on
\gls{ume}, where ratings target pronunciation in read speech. 
WER shows only moderate negative correlations on both datasets. 
The raw-encoder \gls{dtw} baseline is
strong on \gls{ume} ($\rho=-0.50$), but it requires matched recordings of the same text from native and \gls{l2} speakers and is therefore not applicable to \gls{sandi}. 
These baselines provide reference points for interpreting our
native-reference distances, which do not require matched recordings.


\begin{table}[t]
	\centering 	\footnotesize
	\setlength{\tabcolsep}{0.3em}
	\renewcommand{\arraystretch}{1.05}
	\caption{Effect of random boundary shift in milliseconds~(ms) for different \pc es and averaging strategies~(Avg.) using the \gls{lbs} native corpus.
    Values are Spearman $\rho$ of the cosine distance on the \gls{sandi} Train, averaged over three random seeds.
    }
	\label{tab:jitter}
	\begin{tabular}{|l|l|cccccc|}
		\hline
		\multirow{2}{*}{\textbf{Class}} &		\multirow{2}{*}{ \textbf{Avg.}}
		& \multicolumn{6}{c|}{\textbf{Random Boundary Shift [ms]}} \\
		\cline{3-8}
		& &\bf 0 & \bf20 & \bf40& \bf60 & \bf80 & \bf120 \\
		\hline \hline
		\multirow{2}{*}{Diphone}
		& Center
		& $-0.46$
		& $-0.45$
		& $-0.42$
		& $-0.36$
		& $-0.30$
		& $-0.17$ \\ \cline{2-8}
		
		& Full
		& $-0.51$
		& $-0.50$
		& $-0.47$
		& $-0.41$
		& $-0.33$
		& $-0.19$ \\ \hline
		
		\multirow{2}{*}{Triphone}
		& Center
		& $-0.55$
		& $-0.55$
		& $-0.52$
		& $-0.48$
		& $-0.42$
		& $-0.29$ \\ \cline{2-8}
		
		& Full
		& $\mathbf{-0.56}$
		& $\mathbf{-0.56}$
		& $\mathbf{-0.54}$
		& $\mathbf{-0.51}$
		& $\mathbf{-0.46}$
		& $\mathbf{-0.34}$ \\
		\hline
	\end{tabular}
\end{table}

\begin{table}[t]
	\centering
	\footnotesize
	\setlength{\tabcolsep}{0.55em}
	\renewcommand{\arraystretch}{1.05}
	\caption{Spearman $\rho$ for three native-reference~(Native-Ref) corpora evaluated on \gls{sandi} Train and \gls{ume} learner subset using three distance metrics: Euclidean (Euc), Mahalanobis (Maha), and cosine distance.
    \gls{ume} native is held out for testing, see \cref{tab:correlation_template}.}
	\label{tab:codebooks_sandi_ume}
	\begin{tabular}{|c|c|c|c||c|c|c|}
		\hline
		\bf Native-Ref & \multicolumn{3}{c||}{\textbf{\gls{sandi} Train}}
		& \multicolumn{3}{c|}{\textbf{\glsentryshort{ume}}} \\
		\cline{2-7}
		\bf Corpus & \textbf{Euc} & \textbf{Maha} & \textbf{Cosine}
		& \textbf{Euc} & \textbf{Maha} & \textbf{Cosine} \\
		\hline
		\textbf{TIMIT}
		& $-0.44$ & $-0.43$ & $-0.51$ & $-0.25$ & $-0.26$ & $-0.13$ \\
		\hline
		\textbf{\gls{lbs}}
		& $-0.52$ & $-0.51$ & $\bf -0.56$
		& $-0.27$ & $\bf -0.28$ & $-0.13$ \\
		\hline
		\textbf{\glsentryshort{ume}}
		& $-0.35$ & $-0.32$ & $-0.48$ & \multicolumn{3}{c|}{$-$} \\
		\hline
	\end{tabular}
    \vspace{-0.3cm}
\end{table}
\begin{table}[t]
	\centering
	\footnotesize
	\setlength{\tabcolsep}{1.2em}
	\renewcommand{\arraystretch}{1.05}
	\caption{
		Spearman's $\rho$ between distances and \gls{sandi} Train proficiency scores or \gls{ume} learner pronunciation scores. 
        We report Euclidean (Euc), Mahalanobis
		(Maha), and cosine distance metrics under three native-reference coordinate
		constructions: full (no projection), random~(Rnd.) and \gls{svd} projections, both of
		rank $32$. All experiments use triphone classes with \fu{} averaging.
		More negative values indicate better agreement with proficiency.
		The best correlation in each row is highlighted.
	}
	\label{tab:spearman}
	\begin{tabular}{|c|c||c|c|c|}
		\hline
		\multirow{2}{*}{\textbf{\gls{l2} Task}} & \multirow{1}{*}{\textbf{Distance}} & \multicolumn{3}{c|}{\textbf{Spearman $\rho$}} \\
		\cline{3-5}
		&
        \bf Measure
		& \textbf{Full} & \textbf{Rnd.} & \textbf{\gls{svd}} \\
		\hline
		\multirow{3}{*}{\textbf{\gls{sandi} Train}}
		& Euc
		& $-0.50$ & $-0.46$ & $\bf -0.52$ \\ 
		\cline{2-5}
		& Maha
		& $-0.44$ & $-0.46$ & $\bf -0.51$ \\
		\cline{2-5}
		& Cosine
		& $\bf -0.58$ & $-0.53$ & $-0.56$ \\
		\hline \hline
		\multirow{3}{*}{\textbf{\glsentryshort{ume}}}
		& Euc
		& $\bf -0.28$ & $-0.26$ & $-0.27$ \\ 
		\cline{2-5}
		& Maha
		& $\bf -0.29$ & $-0.26$ & $-0.28$ \\
		\cline{2-5}
		& Cosine
		& $\bf -0.19$ & $-0.11$ & $-0.13$ \\
		\hline
	\end{tabular}
\end{table}

\subsection{Effect of Phone Alignment}
\label{subsec:align}
Our proposed distances do not rely on likelihoods or posterior probabilities from an acoustic model, nor on \gls{gop} scores.
However, the computation of the \pc{} centroids still depends on phone-level segmentations.
Therefore, we examine the robustness to random phone-boundary perturbations and verify whether the distance-proficiency correlation is preserved across different \pc{} types and averaging strategies.
When computing the \gls{l2} \pc{} centroids, we randomly perturb the phone boundaries by shifting them with zero-mean Gaussian noise of varying standard deviation, while keeping the phone sequence fixed.
We evaluate diphone and triphone \pc es with \cu{} and \fu{} averaging on \gls{sandi} Train, and calculate the distances with respect to \pc{} coordinates obtained from \gls{lbs} native-reference corpus. 
As shown in \cref{tab:jitter}, increasing the magnitude of the boundary perturbation generally weakens the negative association with proficiency, although its effect depends on the \pc{} type and averaging strategy. 
Triphone representations are more robust than diphones, and \fu{} averaging retains stronger negative correlations than \cu{} averaging as the perturbation increases. 
In particular, the two triphone strategies are nearly identical without perturbation, whereas \fu{} averaging remains more robust under larger boundary perturbations.

\subsection{Different Coordinate Constructions}

\Cref{tab:codebooks_sandi_ume} shows that the proposed distances are consistently negatively correlated with proficiency across all native-reference and \gls{l2} evaluation corpora.
Changing the native-reference corpus has only a limited effect on the correlations, whereas the overall correlation level differs between the two \gls{l2} corpora.
For both tasks, \gls{lbs} native-reference corpus turns out to be the best choice.
On \gls{sandi}, distances calculated with cosine distance reach $\rho=-0.56$, while on \gls{ume} the strongest result ($\rho=-0.28$) is obtained with diagonal Mahalanobis distance.
The weaker correlation on \gls{ume} is expected, since each recording contains only a short read sentence, leading to fewer observed \pc es and possibly noisier utterance-level centroids. 
This consistency indicates that the effect is not tied to a
single native-reference corpus.

\Cref{tab:spearman} compares the proposed \gls {svd}-based L2 coordinates with the full feature space, using the original centroid space $\text{\mnative}$.
By the Johnson-Lindenstrauss projection lemma~\cite{johnson1984extensions}, random projections should approximately preserve distances.
The negative random-projection results serve as a sanity check that proficiency-related information is present in the proposed geometry. 
However, \gls{svd} is generally stronger than random
projection and is close to the full space: it is best for Euclidean and
Mahalanobis distances on \gls{sandi}, and within $0.01$ of the full space for
Euclidean and Mahalanobis distances on \gls{ume}. 
Therefore, the \gls{svd} subspace provides a compact native-reference coordinate system that preserves the relevant acoustic and phonetic variation while reducing computation from 2520 to 8 seconds, corresponding to an approximately $ 315$-times speedup.

As additional controls, replacing the self-supervised features with MFCCs
substantially weakened the correlations, whereas constructing the native
reference from a single TIMIT dialect group had only a minor effect.


\begin{table}[t]
	\centering
	\footnotesize
	\setlength{\tabcolsep}{0.13em}
	\renewcommand{\arraystretch}{1.05}
	\caption{
		Final evaluation on \gls{sandi} Dev and \gls{ume} learner using \gls{lbs} and \gls{ume} native-reference~(Native-Ref) corpora. We report the range of
		matched \pc es $|\mathcal{P}^{\star}|$, Pearson's $r$, raw and partial Spearman's
		$\rho$, for $|\mathcal{P}^{\star}|$ and
		silence ratio as control variables, and the $95\%$ bootstrap confidence interval~(Conf. Int.) for raw $\rho$.
		More negative values indicate stronger agreement with proficiency.
	}
	\label{tab:correlation_template}
	\begin{tabular}{|c|c||c|c||c|c|c|c|c|}
		\hline
		\multirow{4}{*}{\textbf{\gls{l2} Task}}
		& \multirow{3}{*}{\textbf{Native-Ref}}
		& \multicolumn{2}{c||}{\textbf{$\boldmath{| \boldsymbol{\mathcal{P}}^{\star}|}$}}
		& \multicolumn{5}{c|}{\textbf{Correlation}} \\
		\cline{3-9}
		&\multirow{3}{*}{\textbf{Corpus}} &  \multirow{3}{*}{\textbf{min}} &  \multirow{3}{*}{\textbf{max}}
		& \multirow{3}{*}{\textbf{$\mathbf{r}$}}
		& \multicolumn{4}{c|}{\textbf{$\boldsymbol{\rho}$}} \\
		\cline{6-9}
		& & & & & \textbf{Raw} & \textbf{Partial} & \multicolumn{2}{c|}{\textbf{Conf.\ Int.}} \\
		\cline{8-9}
		& & & & & & & \textbf{min} & \textbf{max} \\
		\hline
		\textbf{\gls{sandi} Dev} & \textbf{\gls{lbs}}
		& $9$ & $394$ & $-0.53$ & $-0.53$ & $-0.19$ & $-0.56$ & $-0.49$ \\
		\hline
		\textbf{\gls{ume}} & \textbf{\gls{ume}}
		& $2$ & $54$ & $-0.37$ & $-0.34$ & $-0.27$ & $-0.37$ & $-0.31$ \\
		\hline
	\end{tabular}
    \vspace{-0.4cm}
\end{table}

\subsection{Final Results and Discussion}
\label{subsec:resultsdiscuss}

In \cref{tab:correlation_template}, we report both Pearson $r$ and Spearman $\rho$ correlation coefficients for a final evaluation using the best configuration described in \cref{subsec:setting}. 
Given the high correlations of matched \pc coverage and silence ratio with the proficiency and pronunciation scores, shown in \cref{tab:non_acoustic_baselines,tab:acoustic_baselines}, we additionally report Spearman correlations between distance and the scores while controlling for these variables.
This is known as \emph{partial correlation}, which measures the association between two variables after accounting for the effect of one or more control variables~\cite{cohen2003applied}.
On \gls{sandi} Dev, the native-reference distance shows a clear
negative association with proficiency scores ($r=-0.53$, $\rho=-0.53$).
 After controlling for matched \pc es and silence ratio, the correlation is reduced to $\rho=-0.19$ but remains negative, suggesting that the distance implicitly encodes these factors while also capturing additional variation.
 On \gls{ume}, despite much shorter read utterances and fewer matched \pc es, the distance still correlates with pronunciation rating scores ($r=-0.37$, $\rho=-0.34$, partial $\rho=-0.27$). 
For both datasets, all reported correlations are statistically significant, with their 95\% bootstrap confidence intervals lying entirely below zero. 
As a reference, listener-rated comprehensibility shows only moderate correlations with overall oral proficiency ($\rho=-0.260$ and $-0.345$)~\cite{tergujeff_2021}. 
Therefore, $|\rho|\approx0.3$ is meaningful for human-rating associations, and $|\rho|\approx0.5$ is strong for our proposed method.
In contrast, supervised \gls{sandi} systems reach about $\rho=0.83$~\cite{qian2025speak}, but are trained on proficiency labels and can use information beyond pronunciation.
We used the \gls{ume} task as a complementary evaluation condition to examine whether the proposed native-reference \pc{} geometry transfers to a read-aloud setting.
However, \gls{ume} short sentences provide fewer \pc{} observations per utterance, which may make the utterance-level centroids less stable and the resulting distances less representative.
Moreover, each learner utterance is rated by up to five raters, with an average pairwise human-human Spearman correlation of $\approx0.50$. 
Nevertheless, these results support that our proposed method captures information related to proficiency and pronunciation in spontaneous and read-aloud settings.

\subsection{Limitations}
\label{subsec:resultsdiscuss}
In \gls{sandi}, we evaluate spontaneous speech using holistic proficiency ratings. 
However, pronunciation, fluency, and phonetic coverage are interrelated, making it difficult to attribute the observed association specifically to pronunciation. 
We partly address this limitation by also evaluating on \gls{ume}, where the human ratings specifically target pronunciation quality, although the speech is read aloud. The contribution of other speech-related factors is also reflected in the partial-correlation analysis in \cref{tab:correlation_template}, where controlling for matched \pc{} coverage and silence ratio substantially reduces, but does not eliminate, the negative association.
There is also no theoretical guarantee that the \pc{} coordinates contain only phonetic information, as the underlying representations may encode other factors, such as speaker-dependent characteristics (e.g., accent) or recording conditions. 
Finally, although the proposed distance does not use acoustic-model likelihoods or posterior probabilities, it still requires phone-level segmentation obtained through forced alignment and therefore \gls{asr} transcripts.

\section{Conclusions and Future Work}
\label{sec:conclude}

In this work, we proposed a native-reference \pc{} geometry for measuring \gls{l2} pronunciation deviation using self-supervised speech representations.
By projecting the utterance-level \gls{l2} \pc{} coordinates into a native-reference subspace, we computed distances between matched \pc es and their corresponding native coordinates.
We then showed that the proposed distances have a consistent negative correlation with speakers' proficiency in \gls{sandi} and with learners' pronunciation in the \gls{ume} task, yielding Spearman's $\rho$ of $-0.53$ and $-0.34$, respectively.
The evaluations covered three native-reference corpora, spontaneous and read-aloud \gls{l2} tasks, and comparisons between different native-reference projection spaces.
Using partial correlation analysis and confidence intervals, we confirmed that the negative association between the proposed distances and the \gls{l2} proficiency level is statistically supported.
Unlike other methods, our proposed approach does not require pronunciation labels, read-aloud prompts, or matched native and \gls{l2} recordings of the same text.
Even though our method relies on phone segmentations, we showed that under the best configuration, the correlation remains reasonably robust to random perturbation of the phone boundaries up to $60$ ms.
Future work will investigate how the proposed distances can complement the current state-of-the-art assessment systems as an additional pronunciation-related feature.

\section{Acknowledgments}
\label{sec:acknowlege}
This work was supported by the Strategic Research Council grant numbers 355587, 365233, and 373228.
The authors also acknowledge Aalto University’s Science-IT for providing computational resources.
We thank Zirui Li for helpful discussions related to this work.
OpenAI GPT and Anthropic Claude were used for text reformulation and code development. All outputs were reviewed and verified by the authors.

\vfill\pagebreak
\bibliographystyle{IEEEbib}
\footnotesize{
\bibliography{refs}}

@string{ACL = "Proc.\ ACL"}

@string{ASRU = "Proc.\ IEEE ASRU"}

@string{FindACL = "Findings of ACL"}

@string{ICASSP = "Proc.\ IEEE ICASSP"}

@string{INTERSPEECH = "Proc.\ Interspeech"}

@string{JSTSP = "IEEE Journal of Selected Topics in Signal Processing"}

@string{LREC = "Proc. LREC"}

@string{SLATE = "Proc.\ SLaTE"}

@string{TACL = "Transactions of the Association for Computational Linguistics"}

@article{eckart1936approximation,
	title={The approximation of one matrix by another of lower rank},
	author={Eckart, Carl and Young, Gale},
	journal={Psychometrika},
	volume={1},
	number={3},
	pages={211--218},
	year={1936},
	publisher={Cambridge University Press}
}

@article{raissi2026native,
	title  = {{A Native-Reference Coordinate Geometry for L2 Pronunciation Deviation Using Self-Supervised Speech Models }},
	author = {Raissi, Tina and Phan, Nhan and Kurimo, Mikko},
	journal   = {arXiv:2609.28060},
    year={2026},
}

@book{cohen2003applied,
  title     = {{Applied Multiple Regression/Correlation Analysis for the Behavioral Sciences}},
  author    = {Cohen, Jacob and Cohen, Patricia and West, Stephen G. and Aiken, Leona S.},
  edition   = {3},
  year      = {2003},
  publisher = {Lawrence Erlbaum Associates},
  address   = {Mahwah, NJ}
}

@phdthesis{bluche2015phd,
	title={Deep Neural Networks for Large Vocabulary Handwritten Text Recognition},
	author={Bluche, Théodore},
	year={2015},
	school={LIMSI-CNRS, Lyon, France},
	href={http://www.tbluche.com/phd.html}
}

@misc{raissi2026code,
  author       = {Raissi, Tina and Phan, Nhan and Wang, Chenxiao and Kurimo, Mikko},
  title        = {{Native-L2-Coordinate-Geometry}},
  year         = {2026},
  howpublished = {\url{https://github.com/aalto-speech/native-L2-coordinate-geometry}}
}

@article{johnson1984extensions,
	title={{Extensions of Lipschitz Mappings into a Hilbert Space}},
	author={Johnson, William B and Lindenstrauss, Joram and others},
	journal={Contemporary Mathematics},
	volume={26},
	number={189-206},
	pages={1},
	year={1984}
}

@inproceedings{bredin2020pyannote,
	title={{Pyannote.audio: Neural Building Blocks for Speaker Diarization}},
	author={Bredin, Herv{\'e} and Yin, Ruiqing and Coria, Juan Manuel and Gelly, Gregory and Korshunov, Pavel and Lavechin, Marvin and Fustes, Diego and Titeux, Hadrien and Bouaziz, Wassim and Gill, Marie-Philippe},
	booktitle=ICASSP,
	pages={7124--7128},
	year={2020}
}

@misc{silero2021vad,
	title={Silero {VAD}: Pre-trained enterprise-grade voice activity detector},
	author={{Silero Team}},
	howpublished={\url{https://github.com/snakers4/silero-vad}},
	year={2021}
}

@article{xu2026fireredasr2s,
	title={{FireRedASR2S: A State-of-the-Art Industrial-Grade All-in-One Automatic Speech Recognition System}},
	author={Xu, Kaituo and Jia, Yan and Huang, Kai and Chen, Junjie and Li, Wenpeng and Liu, Kun and Xie, Feng-Long and Tang, Xu and Hu, Yao},
	journal={arXiv:2603.10420},
	year={2026}
}

@inproceedings{minematsu2002english,
	title={{English Speech Database Read by Japanese Learners for CALL System Development}},
	author={Minematsu, Nobuaki and Tomiyama, Yoshihiro and Yoshimoto, Kei and Shimizu, Katsumasa and Nakagawa, Seiichi and Dantsuji, Masatake and Makino, Shozo},
	booktitle={LREC},
	year={2002}
}

@inproceedings{povey2015librispeech,
	author={Panayotov, V. and Chen, G. and Povey, D. and Khudanpur, S.},
	title={{LibriSpeech: An ASR Corpus Based on Public Domain Audio Books}},
	booktitle=ICASSP,
	year=2015,
}

@article{tergujeff_2021,
	title = {{Second Language Comprehensibility and Accentedness across Oral Proficiency Levels: {{A}} Comparison of Two {{L1s}}}},
	shorttitle = {Second Language Comprehensibility and Accentedness across Oral Proficiency Levels},
	author = {Tergujeff, Elina},
	year = 2021,
	month = aug,
	journal = {System},
	volume = {100},
	pages = {102567},
}

@article{chernyak2024perceptual,
	title={A perceptual similarity space for speech based on self-supervised speech representations},
	author={Chernyak, Bronya R and Bradlow, Ann R and Keshet, Joseph and Goldrick, Matthew},
	journal={The Journal of the Acoustical Society of America},
	volume={155},
	number={6},
	pages={3915--3929},
	year={2024},
	publisher={Acoustical Society of America}
}

@article{bartelds2020new,
	title={A new acoustic-based pronunciation distance measure},
	author={Bartelds, Martijn and Richter, Caitlin and Liberman, Mark and Wieling, Martijn},
	journal={Frontiers in Artificial Intelligence},
	volume={3},
	pages={39},
	year={2020},
	publisher={Frontiers Media SA}
}

@inproceedings{parcollet25_interspeech,
	title     = {{Loquacious Set: 25,000 Hours of Transcribed and Diverse English Speech Recognition Data for Research and Commercial Use}},
	author    = {Titouan Parcollet and Yuan Tseng and Shucong Zhang and Rogier C. {van Dalen}},
	year      = {2025},
	booktitle = INTERSPEECH,
	pages     = {4053--4057},
}

@inproceedings{raissi24_interspeech,
	title     = {{Investigating the Effect of Label Topology and Training Criterion on ASR Performance and Alignment Quality}},
	author    = {Tina Raissi and Christoph Lüscher and Simon Berger and Ralf Schlüter and Hermann Ney},
	year      = {2024},
	booktitle = INTERSPEECH,
	pages     = {3899--3903},
}

@article{zelasko2021lhotse,
	title={{Lhotse: a Speech Data Representation Library for The Modern Deep Learning Ecosystem}},
	author={{\.Z}elasko, Piotr and Povey, Daniel and Trmal, Jan and Khudanpur, Sanjeev and others},
	journal={arXiv:2110.12561},
	year={2021}
}

@inproceedings{rybach2011rasr,
	title={{RASR-The RWTH Aachen University Open Source Speech Recognition Toolkit}},
	author={Rybach, David and Hahn, Stefan and Lehnen, Patrick and Nolden, David and Sundermeyer, Martin and T{\"u}ske, Zoltan and Wiesler, Simon and Schl{\"u}ter, Ralf and Ney, Hermann},
	booktitle=ASRU,
	year={2011}
}

@article{garofolo1993timit,
	title={{TIMIT Acoustic-Phonetic Continuous Speech Corpus}},
	author={Garofolo, John S and Lamel, Lori F and Fisher, William M and Pallett, David S and Dahlgren, Nancy L and Zue, Victor and Fiscus, Jonathan G},
	year={1993},
	publisher={Linguistic Data Consortium}
}

@article{mcintosh2026self,
	title={{Self-Supervised} {Speech} {Comparison} for {L2} {Phone}, {Rhythm}, and {Intonation} {Scoring}},
	author={McIntosh, Stephen and Smit, Reuben and Saito, Daisuke and Minematsu, Nobuaki and Kamper, Herman},
	journal={arXiv:2607.13721},
	year={2026}
}

@article{chen2022wavlm,
	title={{WavLM}: {Large-Scale} {Self-Supervised} {Pre-Training} for {Full} {Stack} {Speech} {Processing}},
	author={Chen, Sanyuan and Wang, Chengyi and Chen, Zhengyang and Wu, Yu and Liu, Shujie and Chen, Zhuo and Li, Jinyu and Kanda, Naoyuki and Yoshioka, Takuya and Xiao, Xiong and others},
	journal=JSTSP,
	volume={16},
	number={6},
	pages={1505--1518},
	year={2022},
}

@inproceedings{choi2026b,
	title={[b]=[d]-[t]+[p]: {Self-Supervised} {Speech} {Models} {Discover} {Phonological} {Vector} {Arithmetic}},
	author={Choi, Kwanghee and Yeo, Eunjung and Cho, Cheol Jun and Harwath, David and Mortensen, David R},
	booktitle=FindACL,
	pages={11048--11069},
	year={2026}
}

@inproceedings{choi2025leveraging,
	title={{Leveraging} {Allophony} in {Self-Supervised} {Speech} {Models} for {Atypical} {Pronunciation} {Assessment}},
	author={Choi, Kwanghee and Yeo, Eunjung and Chang, Kalvin and Watanabe, Shinji and Mortensen, David R},
	booktitle=ACL,
	pages={2613--2628},
	year={2025}
}

@inproceedings{pasad2023comparative,
	title={{Comparative} {Layer-Wise} {Analysis} of {Self-Supervised} {Speech} {Models}},
	author={Pasad, Ankita and Shi, Bowen and Livescu, Karen},
	booktitle=ICASSP,
	year={2023},
}

@inproceedings{yang21c_interspeech,
	title={{SUPERB}: {Speech} {Processing} {Universal} {PERformance} {Benchmark}},
	author={Shu-wen Yang and Po-Han Chi and Yung-Sung Chuang and Cheng-I Jeff Lai and Kushal Lakhotia and Yist Y. Lin and Andy T. Liu and Jiatong Shi and Xuankai Chang and Guan-Ting Lin and Tzu-Hsien Huang and Wei-Cheng Tseng and Ko-tik Lee and Da-Rong Liu and Zili Huang and Shuyan Dong and Shang-Wen Li and Shinji Watanabe and Abdelrahman Mohamed and Hung-yi Lee},
	year={2021},
	booktitle=INTERSPEECH,
	pages={1194--1198},
}

@inproceedings{wells2022phonetic,
	title={{Phonetic} {Analysis} of {Self-Supervised} {Representations} of {English} {Speech}},
	author={Wells, Dan and Tang, Hao and Richmond, Korin},
	booktitle=INTERSPEECH,
	pages={3583--3587},
	year={2022}
}

@article{pasad2024self,
	title={{What Do} {Self-Supervised} {Speech} {Models} {Know About} {Words}?},
	author={Pasad, Ankita and Chien, Chung-Ming and Settle, Shane and Livescu, Karen},
	journal=TACL,
	volume={12},
	pages={372--391},
	year={2024}
}

@inproceedings{pasad2021,
	author={Pasad, Ankita and Chou, Ju-Chieh and Livescu, Karen},
	booktitle=ASRU,
	title={{Layer-Wise} {Analysis} of a {Self-Supervised} {Speech} {Representation} {Model}},
	year={2021},
	volume={},
	number={},
	pages={914--921},
}

@inproceedings{qian2025speak,
	title={Speak \& {Improve} {Challenge} 2025},
	author={Qian, Mengjie and Knill, Kate M and Bann{\`o}, Stefano and Tang, Siyuan and Karanasou, Penny and Gales, Mark JF and Nicholls, Diane},
	booktitle=SLATE,
	pages={41--45},
	year={2025}
}

@inproceedings{knill25_slate,
	title     = {{Introducing the Speak \& Improve Corpus 2025: an L2 English Speech Corpus for Language Assessment and Feedback}},
	author    = {Kate M. Knill and Diane Nicholls and Mark J.F. Gales and Mengjie Qian and Pawel Stroinski},
	year      = {2025},
	booktitle = SLATE,
	pages     = {167--171},
}

@article{spearman_1904,
	title={The {Proof} and {Measurement} of {Association} Between {Two} {Things}},
	volume={15},
	journal=AJP,
	author={Spearman, Charles},
	month=jan,
	year={1904},
	pages={72},
}

@article{pearson_1896,
	title={{VII}. {Mathematical} {Contributions} to the {Theory} of {Evolution}.—{III}. {Regression}, {Heredity}, and {Panmixia}},
	number={187},
	urldate={2026-09-05},
	journal={Philosophical Transactions of the Royal Society of London},
	author={Pearson, Karl},
	month=dec,
	year={1896},
	pages={253--318},
}

@article{al-ghezi_ASA2023a,
	title = {Automatic {Speaking} {Assessment} of {Spontaneous} {L2} {Finnish} and {Swedish}},
	volume = {20},
	doi = {10.1080/15434303.2023.2292265},
	language = {en},
	number = {4-5},
	urldate = {2025-03-11},
	journal = {Language Assessment Quarterly},
	author = {Al-Ghezi, Ragheb and Voskoboinik, Katja and Getman, Yaroslav and Von Zansen, Anna and Kallio, Heini and Kurimo, Mikko and Huhta, Ari and Hildén, Raili},
	month = oct,
	year = {2023},
	pages = {421--444},
}

@inproceedings{kyriakopoulos_phone_distrance_2018,
	title = {A {Deep} {Learning} {Approach} to {Assessing} {Non}-native {Pronunciation} of {English} {Using} {Phone} {Distances}},
	booktitle = INTERSPEECH,
	author = {Kyriakopoulos, Konstantinos and Knill, Kate and Gales, Mark},
	year = {2018},
	pages = {1626--1630},
}

@article{chen_speechrater_2018,
    title = {Automated {Scoring} of {Nonnative} {Speech} {Using} the \textit{{SpeechRater}}$^{\textrm{{SM}}}$ v. 5.0 {Engine}},
    volume = {2018},
    issn = {2330-8516, 2330-8516},
    url = {https://onlinelibrary.wiley.com/doi/10.1002/ets2.12198},
    doi = {10.1002/ets2.12198},
    language = {en},
    number = {1},
    urldate = {2026-09-08},
    journal = {ETS Research Report Series},
    author = {Chen, Lei and Zechner, Klaus and Yoon, Su‐Youn and Evanini, Keelan and Wang, Xinhao and Loukina, Anastassia and Tao, Jidong and Davis, Lawrence and Lee, Chong Min and Ma, Min and Mundkowsky, Robert and Lu, Chi and Leong, Chee Wee and Gyawali, Binod},
    month = dec,
    year = {2018},
    pages = {1--31},
}

\end{document}